# From Pixels to Predicates

*Structuring urban perception with scene graphs*

Yunlong Liu[1], Shuyang Li[2,3], Pengyuan Liu[4], Yu Zhang[5] and Rudi Stouffs[6]
[1,5]*School of Architecture, Southeast University, China.*
[2,6]*College of Design and Engineering, National University of Singapore, Singapore.*
[3]*Future Cities Laboratory, Singapore-ETH Centre, Singapore.*
[4]*Division of Urban Studies and Social Policy, University of Glasgow, United Kingdom.*
[1]*lyl_arch@seu.edu.cn, 0009-0001-8392-7420*
[2]*shuyangli@u.nus.edu, 0000-0002-8780-7469*
[4]*pengyuan.liu@glasgow.ac.uk, 0000-0002-5443-5910*
[5]*zhangyuseu@seu.edu.cn*
[6]*stouffs@nus.edu.sg, 0000-0002-4200-5833*

**Abstract.** Perception research is increasingly modelled using streetscapes, yet many approaches still rely on pixel features or object co-occurrence statistics, overlooking the explicit relations that shape human perception. This study proposes a three-stage pipeline that transforms street view imagery (SVI) into structured representations for predicting six perceptual indicators. In the first stage, each image is parsed using an open-set Panoptic Scene Graph model (OpenPSG) to extract object–predicate–object triplets. In the second stage, compact scene-level embeddings are learned through a heterogeneous graph autoencoder (GraphMAE). In the third stage, a neural network predicts perception scores from these embeddings. We evaluate the proposed approach against image-only baselines in terms of accuracy, precision, and cross-city generalization. Results indicate that (i) our approach improves perception prediction accuracy by an average of 26% over baseline models, and (ii) maintains strong generalization performance in cross-city prediction tasks. Additionally, the structured representation clarifies which relational patterns contribute to lower perception scores in urban scenes, such as (graffiti) – [on] – (wall) and (car) – [parked on] – (sidewalk). Overall, this study demonstrates that graph-based structure provides expressive, generalizable, and interpretable signals for modelling urban perception, advancing human-centric and context-aware urban analytics.

## 1. Introduction

Street view imagery (SVI) has become a mainstream data source for urban analytics because it captures façades, signage, vegetation, and other micro-scale cues at eye-level details that are underrepresented in overhead remote sensing and traditional statistics (Biljecki & Ito, 2021). Recent reviews highlight a surge in SVI studies on urban perception, gaining increasing importance in planning, transportation, and public health (Ito et al., 2024).

Early analytical pipelines heavily relied on hand-crafted or pixel-level features (e.g., GIST descriptors, colour, texture) and later adopted deep convolutional neural network (CNN) features to detect or segment visual elements within images. These models typically identify what objects are present and roughly where they are located. Subsequent studies incorporated object counts or semantic segmentation to move beyond raw pixel data (Naik et al., 2014; Dubey et al., 2016; Liu et al., 2023). However, such representations still describe presence rather than relationships. For example, "a car parked on a sidewalk" versus "a car moving along the roadway" conveys different meanings because human judgments depend on interactions among objects and their spatial context, not merely on which objects exist. Reducing both cases to the same categorical set {car, road} omits relational cues that influence perception, prompting a shift from simple co-occurrence statistics toward structured representations that explicitly encode inter-object relationships.

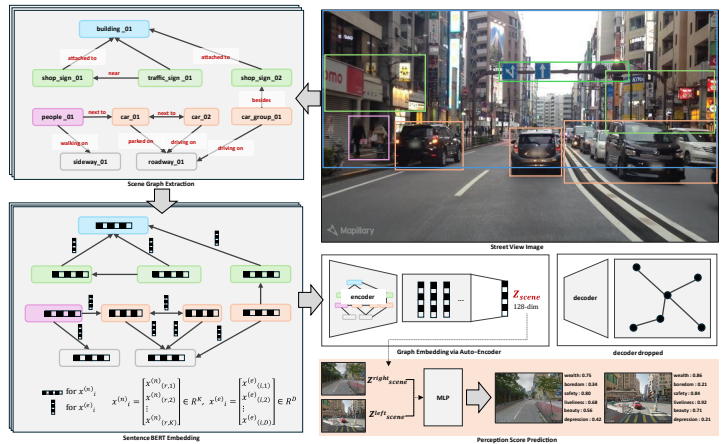

*Figure 1. Research framework overview*

Scene graphs (SGs) provide structure by representing images as object–predicate–object triplets (Johnson et al., 2015). Recent advances extend this paradigm to Panoptic Scene Graphs (PSG), which ground nodes on panoptic segments and jointly predict relations, thereby enhancing object–context coherence in complex scenes (Yang et al., 2022). However, urban environments are exceptionally diverse, and region-specific street furniture, signage, or cleanliness cues often do not appear in predefined label sets of conventional models. Open-vocabulary PSG/SG methods address this limitation by leveraging vision-language learning to expand label sets, allowing the model to recognize new objects and relations beyond the original dataset. This flexibility improves the model's ability to describe and reason about real-world urban scenes (Zhou et al., 2024; Zhang et al., 2023) (see Figure 1). Overall, these developments indicate that open-vocabulary and panoptic scene graphs encode richer semantic and contextual information for urban perception than traditional pixel-based or object-count approaches.

To evaluate whether relation-aware representations improve urban perception prediction, we employ datasets that provide human perception labels and street-level imagery for model training and validation. Specifically, we combine Place Pulse 2.0 (Dubey et al., 2016) and Mapillary Street-Level Sequences (Warburg et al., 2020) to enable both supervised learning on perceptual judgments and cross-city generalization testing under real-world visual conditions.

Place Pulse 2.0 provides a large, multi-city benchmark of crowdsourced pairwise comparisons across six dimensions: safe, lively, boring, wealthy, depressing, and beautiful (Dubey et al., 2016), allowing the model to learn from human evaluations of urban scenes. Mapillary Street-Level Sequences provide extensive, geo-tagged street-level imagery collected under diverse conditions (Warburg et al., 2020). We selected subsets from Tokyo and Amsterdam as independent test domains to assess cross-city generalization, ensuring that the trained model can transfer effectively beyond the cities included in Place Pulse.

The study design and main findings are summarized as follows:

- Proposed pipeline: We develop a pipeline (see Figure 1) that transforms street view imagery into scene graph representations and predicts six perceptual dimensions.

- Performance improvement: Encoding SVI as structured scene graphs yields higher prediction accuracy than image-only baselines (ResNet (He et al., 2016), ViT (Dosovitskiy et al., 2021), CLIP (Radford et al., 2021)) across all perception dimensions, demonstrating that object–relation structures benefit prediction.

- Cross-city generalization: Models trained on Place Pulse 2.0 maintain stable performance on Mapillary subsets, indicating that the graph-structured representations improve cross-city transferability.

## 2. Methodology

### 2.1. RELATIONAL PARSING

The Place Pulse 2.0 dataset is used to train the model, providing supervision across six perceptual dimensions. The dataset is randomly divided into training/validation/test partitions in a 6:3:1 ratio. Each image is then converted into a scene graph following the pipeline illustrated in Figure 2. In this representation, nodes correspond to urban entities, and edges denote semantic relations between them. Object and predicate texts are encoded with Sentence-BERT (Reimers & Gurevych, 2019), a text embedding model, which maps textual descriptions into high-dimensional semantic vectors. This process can be formalized as:

$$x_n = f_{SBERT}(attr_x)$$

where $attr_x$ denotes the textual description of an object or predicate (e.g., "building",

"on", "sidewalk"), and $f_{\text{SBERT}}(\cdot)$ represents the Sentence-BERT encoder that maps each text into a semantic vector $x_n \in R^{d_t}$.

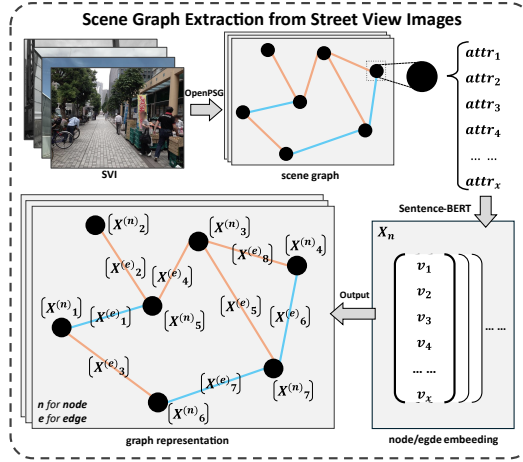

*Figure 2. Scene graph construction and feature encoding*

## 2.2. GRAPH EMBEDDINGS

To learn meaningful representations from scene graphs, we employ GraphMAE, a self-supervised model framework for learning structured graph representations without manually labelled data (Hou et al., 2022). GraphMAE masks portions of the input graph and trains the model to reconstruct the missing information, thereby forcing it to understand the underlying spatial and semantic structure of the urban scene (see Figure 3). After training, the decoder of GraphMAE is discarded and the encoder produces scene-level embeddings (128-dim in our implementation) for downstream urban perception prediction.

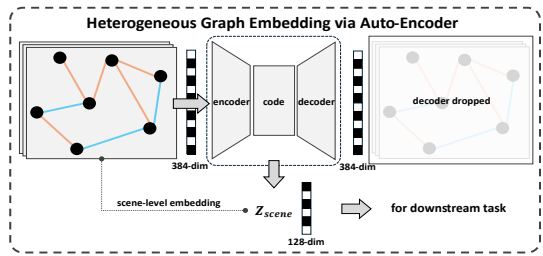

*Figure 3. Masked graph autoencoding for scene-level representation learning*

## 2.3. PERCEPTION INFERENCE

To enable the model to learn from human perceptual judgments, we adopt a pairwise-comparison learning approach as the training mechanism. This approach trains a model to determine, for each pair of images, which scene is perceived more positively on a given perceptual dimension. Classical formulations such as Bradley–Terry and RankNet can translate binary comparisons into continuous perceptual scores (Bradley

# From Pixels to Predicates: Structuring urban perception with scene graphs

& Terry, 1952; Burges et al., 2005). In our implementation, each pair of scene embeddings is passed through a weight-shared multilayer perceptron (MLP) to yield scores, as shown in Figure 4. The probability of one scene being preferred over another is computed as $P(left > right) = \sigma(s_{left} - s_{right})$, and the model is trained via cross-entropy using the true pairwise labels.

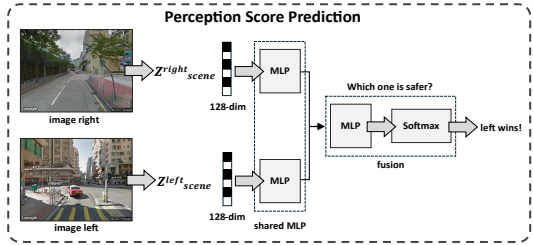

*Figure 4. Pairwise perception prediction framework*

Different backbones are compared for the Bradley–Terry head, including ResNet-50, ViT-B/16, and CLIP ViT-B/32. We evaluate model performance on the Place Pulse 2.0 dataset with accuracy per dimension and mean accuracy. To further assess cross-city generalization, the trained models are tested on Mapillary street-level imagery from Tokyo and Amsterdam.

## 3. Results

The first experiment examines how structured representations improve perception prediction relative to image-based models (ResNet50, ViT-B/16, and CLIP ViT-B/32). All approaches were trained and evaluated within the same dataset, allowing a controlled comparison of representational capacity.

The baseline models shared the same pairwise comparison formulation, implemented with identical data splits and loss functions as the graph-based approach. Each image-only baseline was pretrained on large-scale datasets but trained here with frozen backbones to ensure fair evaluation.

Results show that the graph-based model achieves higher accuracy, precision, and area-under-curve (AUC) scores across all perceptual dimensions (Table 1). The largest gains appear in wealth and liveliness predictions (Table 2), where human perception depends strongly on compositional relationships rather than raw texture or colour.

Table 1. Performance comparison across baseline and graph-based models

| Model | AUC | accuracy | recall | f1 | precision |
|---|---|---|---|---|---|
| CLIP | 0.61 | 0.63 | 0.60 | 0.55 | 0.60 |
| CNN | 0.82 | 0.73 | 0.78 | 0.68 | 0.64 |
| ViT | 0.79 | 0.71 | 0.74 | 0.72 | 0.67 |
| **GraphMAE** | **0.84** | **0.87** | **0.89** | **0.85** | **0.83** |

Table 2. Perception prediction performance across six perceptual dimensions

| Model | beautiful | boring | depressing | lively | safety | wealthy | average |
|---|---|---|---|---|---|---|---|
| CLIP | 0.59 | 0.64 | 0.59 | 0.64 | 0.64 | 0.66 | 0.63 |
| CNN | 0.74 | 0.69 | 0.67 | 0.77 | 0.76 | 0.77 | 0.73 |
| ViT | 0.72 | 0.67 | 0.63 | 0.73 | 0.74 | 0.74 | 0.71 |
| **GraphMAE** | **0.87** | **0.84** | **0.83** | **0.88** | **0.87** | **0.90** | **0.87** |

## 3.1. CROSS-CITY GENERALIZATION TEST

The second experiment investigates whether relational representations generalize across distinct urban contexts. To minimize cultural bias, we selected Tokyo and Amsterdam from the Mapillary dataset, two cities that differ in spatial layout, building typology, and streetscape. The pre-trained graph-based model is directly applied to predict on the manually annotated subsets from Tokyo and Amsterdam. These subsets are annotated using the same pairwise comparison method as the Place Pulse dataset, where annotators select from two street-view images the one that better corresponds to a specific perceptual dimension (e.g., safety, liveliness).

Results reveal a moderate decline in performance when evaluated on the Tokyo–Amsterdam subset compared to the full Place Pulse dataset. Test accuracy decreases by approximately 5.6%, while the AUC shows a 3.5% reduction (Table 3), indicating that the relational embeddings can capture higher-level spatial patterns that remain interpretable and predictive across different cities.

The structured graph-based model exhibits strong cross-city generalization, indicating that relational semantics provide a transferable and robust foundation for urban perception prediction.

Table 3. Cross-city generalization performance on Mapillary subsets (Tokyo and Amsterdam)

| Metric | Place Pulse 2.0 | Tokyo–Amsterdam | Change |
|---|---|---|---|
| Accuracy | 0.84 | 0.79 | −5.6% |
| AUC | 0.87 | 0.84 | −3.5% |

## 3.2. QUALITATIVE ANALYSIS

We further inspect selected examples from the Mapillary subset used in the cross-city generalization experiment. Figure 5 arranges representative scenes from Tokyo and Amsterdam along six perceptual dimensions: safety, liveliness, boredom, wealth, depression, and beauty, ordered from lower to higher predicted scores.

A close look at Figure 5 reveals consistent relational motifs across both Tokyo and Amsterdam. In the higher-scoring columns, scenes tend to exhibit spatial enclosure: pedestrians appear near façades, trees or hedges align with sidewalks, and street elements (lamps, benches) create a human-scale rhythm. These object-relation patterns correspond to perceptions of safety, liveliness, and beauty. In contrast, the lower-scoring columns often display road segments with few adjacent structures or human

From Pixels to Predicates: Structuring urban perception with scene graphs presence, producing a sense of emptiness or exposure. These scenes typically lack relational connectivity, such as people–sitting on–benches or bicycles–parked beside–building, or plants–growing along–walls, which the patterns associate with more favourable perceptual scores. While these patterns are broadly consistent across cities, sampling imbalances remain: Tokyo's subset includes more façade-rich streetscapes, whereas Amsterdam's subset contains more open-road or green-dominant views, which may partially influence the perceptual scores.

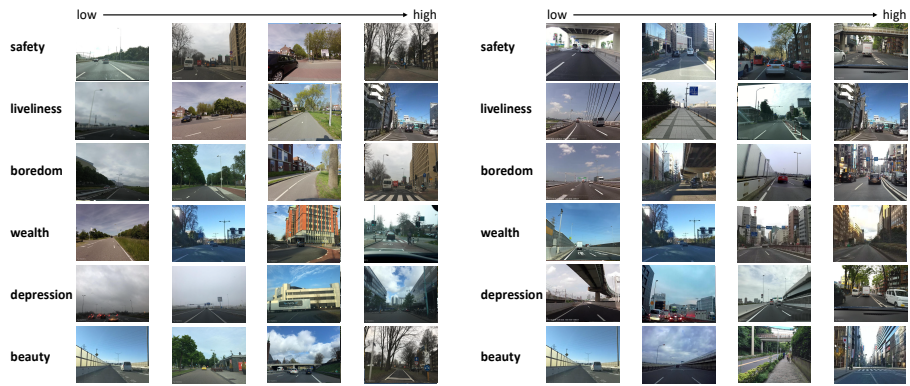

*Figure 5. SVI examples by perception score, Tokyo(right) and Amsterdam(left)*

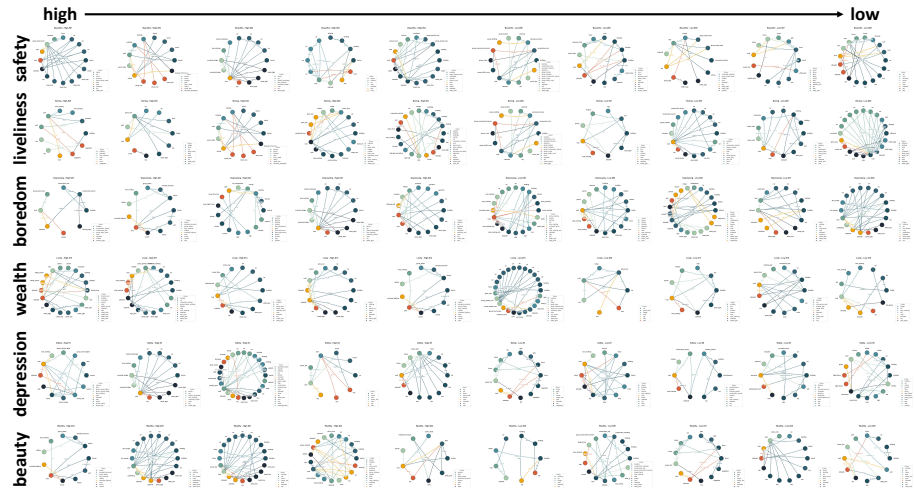

*Figure 6. Scene graph thumbnail overview*

To complement the image grid, Figure 6 visualizes the predicted scene graphs for six perceptual dimensions. Higher-scoring scenes generally display richer relational topology (characterized by more edges) and greater subject diversity (reflected in more node types), suggesting a denser web of interactions among urban elements (see examples in Figure 7). In contrast, some low-scoring cases also contain many nodes and edges but concentrate them within a limited set of repetitive relations—for example, repetitive triplets like car–near–car or road–beside–road. Despite their apparent complexity, these graphs lack the heterogeneity and cross-category links that correlate with positive perceptions. This contrast underscores that variety and cross-type relations, rather than graph density, are more strongly aligned with favourable human perception.

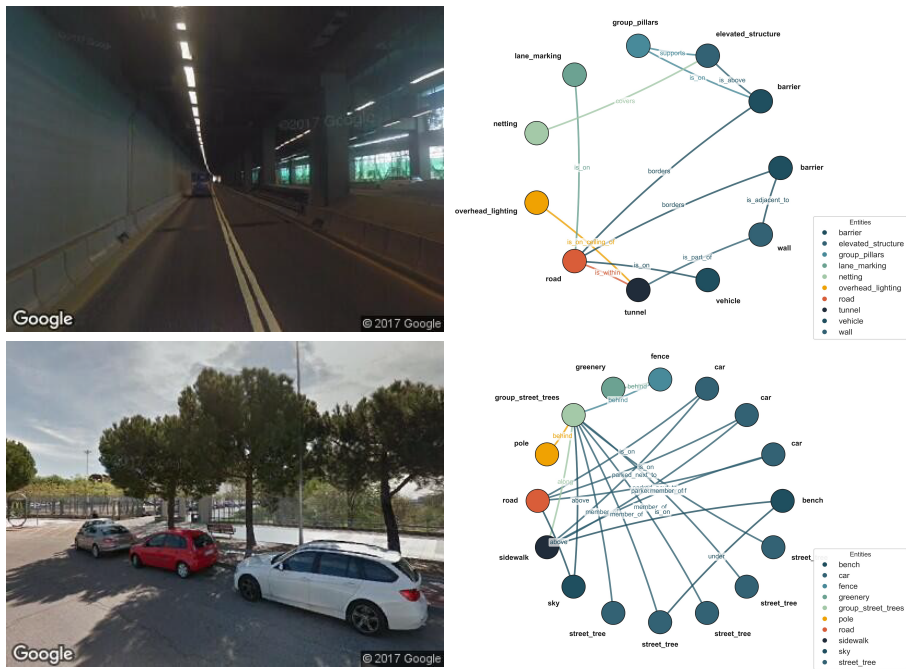

Figure 7. Street views and scene graph examples (top: low beauty score; bottom: high beauty score)

The results suggest that relational structure is a key factor for predicting human perception. Graph-based representations effectively capture spatial and functional interactions, such as proximity between people and buildings, or alignment of vegetation with walkable space, which remain stable across contexts. By embedding these semantics into a deep learning framework, the model bridges the gap between visual appearance and human interpretation, enabling both scalable and interpretable perception analysis.

## 4. Conclusion

This study demonstrates that representing street view imagery through structured scene graphs substantially enhances the modelling of urban perception. Compared with image-only baseline models, the proposed approach consistently achieves higher predictive accuracy across all perceptual dimensions. Moreover, the proposed approach also exhibits cross-city generalization in independent tests conducted in Tokyo and Amsterdam, demonstrating that relational structures captured by scene graphs are transferable across diverse urban contexts.

This approach also holds practical values for evidence-based urban design. The relational patterns identified by the model can serve as diagnostic cues, revealing where perceived discomfort or disorder may arise from spatial misalignment or missing human-environment interactions.

Nonetheless, several limitations point to directions for future research. The cross-city evaluation was based on a limited subset, and expanding it with more representative samples would improve statistical reliability. In addition, the OpenPSG-based scene parsing could be further improved through domain-specific fine-tuning to better capture relational pattern nuances unique to different urban environments.

## Acknowledgements

ChatGPT (OpenAI, 2025) was used to (1) improve writing and (2) assist with coding.